\documentclass[11pt]{article}

\usepackage[preprint]{acl}

\usepackage{times}
\usepackage{latexsym}
\usepackage{amsmath}
\usepackage{amssymb}
\usepackage{url}

\usepackage[T1]{fontenc}

\usepackage[utf8]{inputenc}

\usepackage{microtype}

\usepackage{inconsolata}

\usepackage{graphicx}
\usepackage{array}
\usepackage{booktabs}
\usepackage{dblfloatfix}
\usepackage{xcolor}
\usepackage{fontawesome5}
\usepackage{placeins}
\usepackage{float}

\newcommand{\best}[1]{\textbf{#1}}
\newcommand{\second}[1]{\underline{#1}}
\newcommand{\singletablefont}{\scriptsize}
\definecolor{promptbg}{RGB}{242,242,242}
\definecolor{promptframe}{RGB}{226,226,226}
\definecolor{promptaccent}{RGB}{53,94,141}
\definecolor{prompttitlebg}{RGB}{234,240,247}

\newcommand{\promptpaneltitle}[1]{%
  \vspace{4pt}%
  \noindent\makebox[\columnwidth][c]{%
    \setlength{\fboxsep}{4pt}%
    \colorbox{prompttitlebg}{\small\bfseries\color{promptaccent}#1}%
  }\par
  \vspace{3pt}%
}
\newcommand{\promptbox}[1]{%
  \noindent\makebox[\columnwidth][c]{%
  \setlength{\fboxsep}{4pt}%
  \fcolorbox{promptframe}{promptbg}{%
    \parbox{\dimexpr0.99\columnwidth-2\fboxsep-2\fboxrule\relax}{%
      \raggedright\sloppy\small #1
    }%
  }%
  }%
}

\title{SPAR-Hate: Auditor-Guided Multi-Perspective Role Reasoning for Bilingual Hate Speech Parsing}

\author{Yifan Lyu\textsuperscript{1} \quad Dianqing Lin\textsuperscript{2} \quad Xinran Li\textsuperscript{1} \quad Jiaqi Qiao\textsuperscript{1} \quad Xiujuan Xu\textsuperscript{1,*}\\
\textsuperscript{1}Dalian University of Technology \quad \textsuperscript{2}Inner Mongolia University\\
\texttt{stevelyu811@gmail.com} \quad
\textsuperscript{*}\textbf{Correspondence:}
\href{mailto:xjxu@dlut.edu.cn}{\texttt{xjxu@dlut.edu.cn}}}

\begin{document}
\maketitle
\begin{abstract}
\noindent
\textcolor{red}{\faExclamationTriangle}\enspace
{\color{red}\bfseries
Warning: This paper contains content that may be offensive or harmful.}
\par\medskip
Hate speech research has moved from coarse-grained classification towards structured parsing, where systems jointly identify targets, supporting arguments, and target-level labels. Documents with multiple targets, conflicting local readings, or culturally coded language make these bindings difficult to recover. SPAR-Hate is an auditor-guided multi-perspective role-reasoning framework for bilingual hate speech parsing. It decomposes each document into local focus units, elicits evidence-grounded candidates from Victim, Moderator, and Cultural Bystander perspectives, resolves candidate conflicts under grounding and schema constraints, and reassembles sample-level predictions. Experiments on STATE-ToxiCN and a controlled TBO split show gains across local and API backbones, concentrated on strict joint target--argument--label metrics. Controlled comparisons favour separate candidate generation and arbitration for joint parsing. Within the tested pipeline, role-conditioned prompting yields higher exact-match joint scores than repeated generic prompting, while Chinese overlap-based results depend on the backbone. Structured teacher traces also support training a smaller student model.
\end{abstract}

\section{Introduction}

Hate speech detection identifies hateful expressions whose interpretation depends on context, social judgement, and platform standards~\citep{schmidt-wiegand-2017-survey,10.1145/3232676}. Fine-grained work now includes rationale annotation, target--argument parsing, and tuple extraction~\citep{pavlopoulos-etal-2021-semeval,Mathew2020HateXplainAB,zampieri-etal-2023-target,bai-etal-2025-state}. Structured hate parsing predicts one or more target-level tuples, each linking an attacked target, its supporting argument, and a target-level label.

\begin{figure}[!t]
  \centering
  \includegraphics[width=\columnwidth]{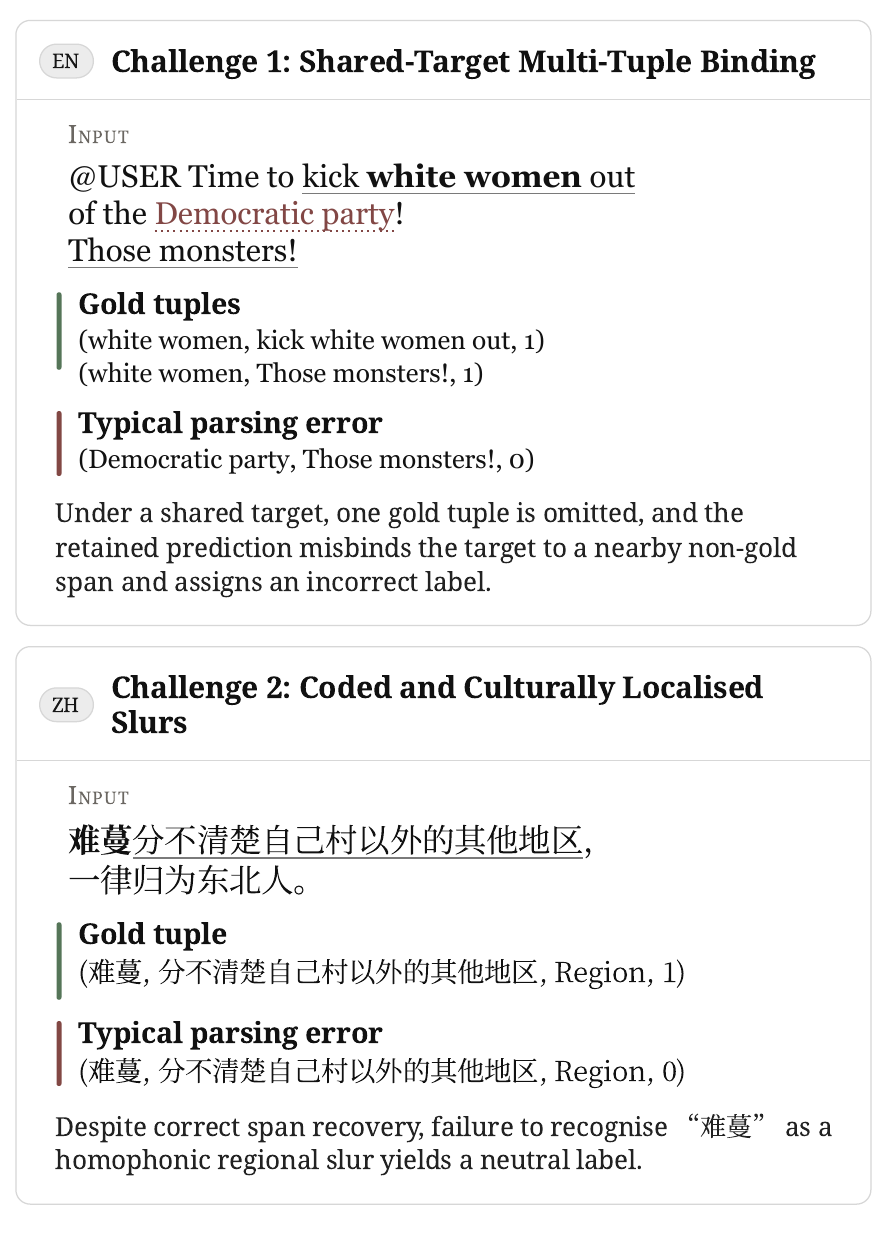}
  \caption{Examples of bilingual parsing errors. The English example shows omitted tuples and incorrect target--argument--label binding; the Chinese example shows a missed culturally localised homophonic slur.}
  \label{fig:intro_challenge}
\end{figure}

Figure~\ref{fig:intro_challenge} shows a shared-target English prediction with an omitted tuple and incorrect target--argument--label binding, alongside a missed Chinese homophonic slur. These failures require consistent tuple prediction and culturally grounded interpretation. Large language models remain skewed towards Western cultural representations~\citep{naous-etal-2024-beer}, weakening performance on implicit, coded, and locally grounded hate~\citep{elsherief-etal-2021-latent,nozza-2021-exposing,ocampo-etal-2023-depth,lin2026exploringcapabilityboundariesllms}. Cultural knowledge can change target identification, evidence selection, and harm attribution. Social position also shapes interpretations of hostile speech~\citep{spears-2021-social-influence}, so the same utterance can support different grounded readings.

SPAR-Hate comprises four stages: Segment, Perspective-Guided Role Generation, Arbitrate, and Reassemble. Phase~1 produces local focus units. Victim, Moderator, and Cultural Bystander generators produce evidence-grounded candidates for each unit. Phase~3 clusters and selects candidates under grounding and schema constraints, and Phase~4 restores sample-level predictions. The Chinese Cultural Bystander receives weak lexical context for coded language and local slang \cite{bai-etal-2025-state,xiao-etal-2024-toxicloakcn,lu-etal-2023-facilitating}.

\paragraph{Contributions.}
The task formulation aligns Chinese and English structured parsing at field level while retaining their original annotation contracts. The framework combines local focus-unit segmentation, three perspective-conditioned generators, constrained arbitration, and sample-level reassembly. Bilingual experiments evaluate joint tuple recovery across local and API backbones. Controlled comparisons examine candidate generation and arbitration, while distillation tests pipeline traces as supervision for a smaller student.

\section{Related Work}
\subsection{From Hate Speech Classification to Structured Tuple Parsing}

Hate speech research began largely as text classification, but class labels alone do not support fine-grained semantic understanding. Early work focused on definitions, features, and classification models \cite{schmidt-wiegand-2017-survey,10.1145/3232676,waseem-hovy-2016-hateful,davidson_automated_2017}. Later multilingual shared tasks and richer annotations showed that multilingual hate analysis requires more than single-label prediction \cite{basile-etal-2019-semeval,ousidhoum-etal-2019-multilingual}. Toxic Spans, HateXplain, TBO, and STATE-ToxiCN then moved the field toward fine-grained localisation, rationale annotation, target-argument parsing, and Chinese quadruple parsing \cite{pavlopoulos-etal-2021-semeval,Mathew2020HateXplainAB,zampieri-etal-2023-target,bai-etal-2025-state}.

SPAR-Hate focuses on local target--argument--label bindings under separate Chinese and English annotation contracts and evaluates their sample-level reconstruction.

\subsection{Implicit Hate, Cultural Context, and Cross-Lingual Fragility}

Implicit, subtle, and context-dependent hate remains difficult to detect. Coded and indirect expressions weaken model performance \cite{elsherief-etal-2021-latent,ocampo-etal-2023-depth,hartvigsen-etal-2022-toxigen}, while HateCheck identifies related functional failures \cite{rottger-etal-2021-hatecheck}. Cross-lingual zero-shot models can also misread language-specific non-hateful taboo expressions as hate signals \cite{nozza-2021-exposing}.

Explainable hate-speech detection uses rationales, social bias frames, and stepwise explanations \cite{Mathew2020HateXplainAB,sap-etal-2020-social,yang-etal-2023-hare}. Structured parsing adds the requirement that target, argument, and label remain jointly aligned. SPAR-Hate supplies cultural knowledge as weak context and retains only text-grounded candidates.

\subsection{Role-Conditioned Reasoning and Constrained Aggregation}

RoleLLM and multi-perspective role-playing show that role conditioning elicits distinct knowledge and reasoning biases \cite{wang-etal-2024-rolellm}. Proposer--aggregator methods coordinate several model outputs \cite{du2023improvingfactualityreasoninglanguage}. AutoGen and MetaGPT organise role allocation and structured workflow handoffs, with MetaGPT encoding these handoffs through standard operating procedures \cite{wu2023autogenenablingnextgenllm,hong2024metagptmetaprogrammingmultiagent}.

SPAR-Hate constrains role outputs to comparable target--argument--label candidates before aggregation. Its distillation traces retain focus-unit decomposition, role hypotheses, conflict diagnosis, and final arbitration, linking the method to chain-of-thought, self-consistency, and reasoning distillation \cite{wei2023chainofthoughtpromptingelicitsreasoning,wang2023selfconsistencyimproveschainthought,shridhar2023distillingreasoningcapabilities,hsieh-etal-2023-distilling}.

\section{Methods: The SPAR-Hate Framework}
\label{sec:methods}

SPAR-Hate combines local focus-unit decomposition, multi-perspective candidate generation, dynamic arbitration, and sample-level reassembly in an auditor-guided framework for bilingual hate parsing. The pipeline breaks document-level parsing into explicit intermediate stages, which helps long texts, multi-target cases, implicit attacks, and culturally coded slang.

\subsection{Task Formulation and Output Contracts}
\label{subsec:task}

Given an input document $D$, the system must predict a set of target-level structured tuples
\begin{equation}
E = \{e_1, e_2, \ldots, e_n\}.
\end{equation}

The Chinese and English benchmarks follow different original annotation contracts. STATE-ToxiCN uses quadruples
\begin{equation}
e_{zh}^{raw} = (target, argument, group, hateful),
\end{equation}
where \texttt{group} denotes the attacked group category. TBO uses triples
\begin{equation}
e_{en}^{raw} = (target, argument, harmful).
\end{equation}
These schemas correspond respectively to the Target--Argument--Hateful--Group annotation contract in STATE-ToxiCN and the target--argument--harmfulness contract in TBO \cite{bai-etal-2025-state,zampieri-etal-2023-target}.

To construct the bilingual main track at field level, the Chinese main track removes \texttt{group} and uses a three-field output
\begin{equation}
e_{zh}^{main} = (target, argument, label).
\end{equation}

For unified notation, each tuple on the aligned bilingual main tracks is written as $e_i = (t_i, a_i, \ell_i)$. Here $t_i$ denotes the attacked target, $a_i$ the attack argument, and $\ell_i \in \{0,1\}$ the target-level harmfulness label. It corresponds to \texttt{harmful} in English and \texttt{hateful} in Chinese. The Chinese four-field extension retains \texttt{group} as an additional field.

The Chinese three-field task supports the bilingual main comparison. A four-field extension retains \texttt{group} and tests adaptation to the original language-specific schema. Separate prompt templates and output constraints preserve both contracts.

\subsection{Framework Overview}
\label{subsec:overview}

\begin{figure*}[!t]
  \centering
  \includegraphics[width=\textwidth]{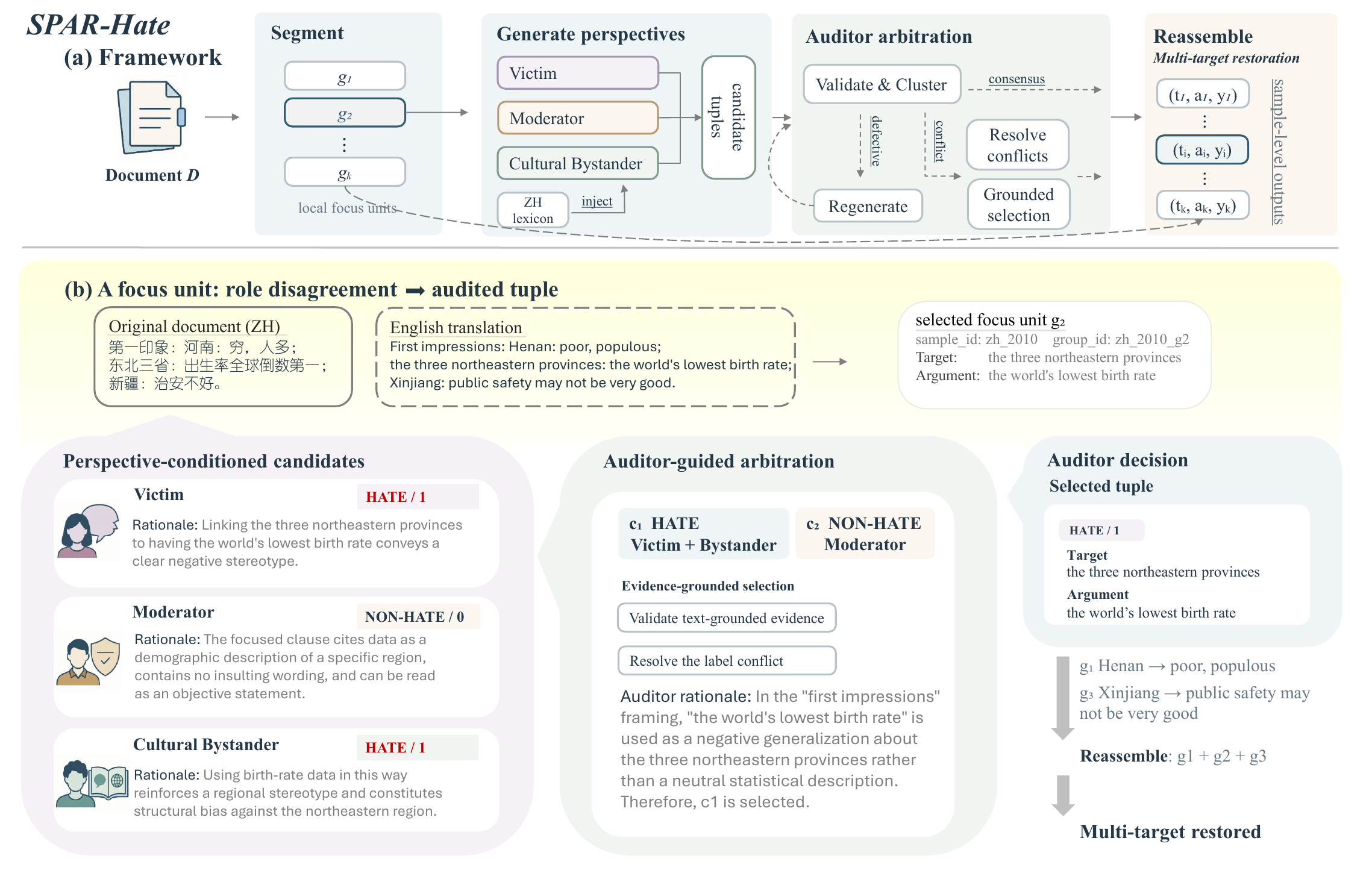}
  \caption{Overview of SPAR-Hate. (a) The pipeline segments bilingual input into local focus units, generates role-conditioned candidates, arbitrates under evidence constraints, and reassembles sample-level outputs. (b) A Chinese focus unit illustrates role disagreement and Auditor selection before multi-target reassembly.}
  \label{fig:spar_overview}
\end{figure*}

SPAR-Hate has four stages: \textit{Segment}, \textit{Perspective-Guided Role Generation}, \textit{Arbitrate}, and \textit{Reassemble}, corresponding to Phase~1--4. The system splits a document into local focus units, generates structured candidates from three perspectives, arbitrates them with a dynamic-beacon procedure, and reassembles benchmark-aligned sample-level outputs. Figure~\ref{fig:spar_overview} presents the full pipeline.

\subsection{Phase 1: Local Focus-Unit Segmentation}
\label{subsec:phase1}

Long texts often contain multiple targets, local stances, and interfering attack fragments. Direct extraction from the full document can therefore produce target--argument mismatches, overly wide arguments, and merged events. Divide-and-conquer frameworks for document-level sentiment parsing suggest the same advantage for structured extraction \cite{wang_danceha_2026}. Phase~1 therefore uses an LLM-based segmenter to map document $D$ to a set of local decision units
\begin{equation}
G = \{g_1, g_2, \ldots, g_k\}.
\end{equation}

Each local unit $g_i$ stores sample-level and local indices (\texttt{group\_id}, \texttt{local\_group\_id}), local text (\texttt{group\_text}), a focus-marked local context (\texttt{focus\_text}), and a soft target anchor (\texttt{canonical\_target}). A focus unit is defined around a target and local intent while retaining enough context for argument grounding. Its boundary need not coincide with a syntactic boundary.

\subsection{Phase 2: Perspective-Guided Role Generation}
\label{subsec:phase2}

Perspective shapes hate judgements \cite{waseem-hovy-2016-hateful,sap-etal-2020-social}. The task-motivated role set covers affected-group harm, platform-governance boundaries, and community interpretation of implicit or culturally coded hate. For each local unit $g_i$, the framework instantiates three role-conditioned generators
\begin{equation}
R = \{r_{\text{victim}}, r_{\text{moderator}}, r_{\text{bystander}}\}.
\end{equation}

Victim emphasises felt harm, exclusion, and microaggressions. Moderator emphasises platform-governance boundaries and explicit rule violations. Cultural Bystander emphasises local cultural context, community slang, pragmatic history, and coded expression. Each role generates one or more structured candidates over the same \texttt{focus\_text}, yielding the role-specific candidate set
\begin{equation}
H_r(g_i) = \{h_r^{(1)}, h_r^{(2)}, \ldots\}.
\end{equation}

SPAR-Hate imposes a strict JSON schema and requires \texttt{argument} to be a supporting substring inside \texttt{focus\_text}. Every judgement includes local textual evidence.

The Chinese Cultural Bystander receives entries from the auxiliary STATE-ToxiCN lexicon. Its 829 \texttt{term/category/definition} records contain no sample identifiers, gold tuples, or instance labels. Retrieved entries provide weak context for the current \texttt{focus\_text} and \texttt{canonical\_target}. Candidates remain subject to focus-span grounding, and lexicon-only singleton candidates are discarded. Lexicon hits occur in 10.1\% of \texttt{ZH-main} groups and 9.6\% of \texttt{ZH-quadruple} groups. Chinese toxicity research motivates this support for slang, homophones, and emoji cloaking \cite{lu-etal-2023-facilitating,bai-etal-2025-state,xiao-etal-2024-toxicloakcn}. Appendix~\ref{sec:appendix_prompts} gives the prompt templates.

\subsection{Phase 3: Local Candidate Clustering and Auditor Arbitration}
\label{subsec:phase3}

Phase~3 filters, clusters, and ranks the role candidates for each local unit. Given
\begin{equation}
\mathcal{H}_i = \bigcup_{r \in R} H_r(g_i),
\end{equation}
the procedure applies quality scoring, cluster formation, triage, and deterministic selection.

\paragraph{Candidate Quality Scoring}
The system first assigns each candidate $c \in \mathcal{H}_i$ a quality score $q(c)$:
\begin{equation}
q(c) = \sum_j \alpha_j \,\phi_j(c).
\end{equation}
Here $\phi_j(c)$ includes grounding, target explicitness, consistency with \texttt{canonical\_target}, structural completeness, label validity, and, in the Chinese four-field setting, \texttt{group}--label coherence. Chinese bystander candidates can receive a small lexicon bonus. Candidates below quality $0.60$ are filtered before clustering. An ungrounded or missing argument fails validation.

\paragraph{Cluster Formation}
Candidates are then clustered by similarity over target, argument, \texttt{harmful}/\texttt{hateful}, and, in the Chinese four-field extension, \texttt{group}. Similarity between a candidate and a cluster is written as
\begin{equation}
\begin{aligned}
\mathrm{sim}(c, C) = \, &\beta_t\, s_t(c, C) + \beta_a\, s_a(c, C) \\
                        &+ \beta_y\, s_y(c, C) + \beta_g\, s_g(c, C),
\end{aligned}
\end{equation}
where $s_t, s_a, s_y, s_g$ denote similarity in \texttt{target}, \texttt{argument}, \texttt{label}, and \texttt{group}. On the three-field main tracks, $\beta_g=0$. The reported field weights are $0.45/0.45/0.05/0.05$, respectively. Each cluster is then compressed into a canonical tuple and assigned a cluster score based on member quality, the number of supporting roles, and any lexicon-aware bonus:
\begin{equation}
\mathrm{Score}(C) = \sum_{c \in C} q(c) + \lambda_1 |\mathrm{supp}(C)| + \lambda_2 \mathbb{I}_{\mathrm{lex}}(C),
\end{equation}
where $\mathrm{supp}(C)$ is the set of roles supporting that cluster. The mainline support-role and lexicon-cluster bonuses are $0.12$ and $0.08$; the candidate-level lexicon bonus is $0.05$.

\paragraph{Triage and Dynamic Soft Beacons}
After clustering, the system routes each local unit into one of three lanes according to role agreement, the margin of the top cluster, and signals of structural failure or missing roles. A clearly dominant top cluster with no obvious defect enters \texttt{consensus}. Missing roles or a lack of valid grounded candidates enters \texttt{defective}. The remaining cases enter \texttt{conflict}. The system then applies grounding-aware, lexicon-aware, and quote-aware soft beacons to adjust cluster ranking. If some roles are marked defective and regeneration is enabled, one bounded regeneration round is allowed. Highly conflicting cases can trigger a second auditor refinement.

\paragraph{Deterministic Fallback and Local Refinement}
Final ranking uses a deterministic cluster-level fallback. The selected primary cluster undergoes local refinement of \texttt{target} and \texttt{argument} boundaries. The reported configuration retains at most two clusters, requires score $\geq 1.05$ and at least two supporting roles, and allows one regeneration round for a defective role. Appendix~\ref{sec:appendix_auditor} reports the complete scoring, routing, and boundedness diagnostics.

\subsection{Phase 4: Sample-Level Multi-Target Reassembly}
\label{subsec:phase4}

In this task setting, local focus-unit decisions are made internally while the benchmark expects sample-level predictions. Phase~4 therefore restores earlier decisions to the output format required by evaluation. The system first aggregates all local tuples by \texttt{sample\_id}, producing
\begin{equation}
\tilde{E}(x)=\biguplus_{g_i \in G(x)} Y(g_i),
\end{equation}
where $Y(g_i)$ is the set of final tuples output by Phase~3 for local unit $g_i$. The main configuration uses a concatenate-first strategy, preserving local provenance and not forcing exact deduplication. The default exported sample-level prediction is therefore
\begin{equation}
\hat{E}(x)=\tilde{E}(x),
\end{equation}
whereas under optional exact deduplication
\begin{equation}
\hat{E}(x)=\mathrm{Dedup}\!\left(\tilde{E}(x)\right).
\end{equation}

The pipeline records focus units, role candidates, conflict diagnoses, and final tuples as teacher traces. Training-split records filtered for argument grounding, tuple validity, and audit confidence provide reasoning-text and JSON supervision for the 4B student.

\section{Experiments}
\label{sec:experiments}

\subsection{Datasets and Evaluation Protocol}
\label{subsec:data_eval}

\paragraph{Datasets}
Experiments are conducted on Chinese STATE-ToxiCN and English TBO \cite{bai-etal-2025-state,zampieri-etal-2023-target}. STATE-ToxiCN keeps its official train/test split. The public release of TBO provides only a test split, so the public test set is deterministically shuffled with \texttt{seed=20260415} and re-divided into 3200/800 train/test subsets. Beyond this repartition, processing is limited to field cleaning, schema normalisation, and split freezing. No extra relabelling is introduced, and multi-target documents are not split at this stage. The paper reports three evaluation tracks, \texttt{ZH-main}, \texttt{EN-main}, and \texttt{ZH-quadruple}, all defined exactly as in Section~\ref{subsec:task}. Dataset composition appears in Table~\ref{tab:data_summary}. TBO numbers are controlled within-paper comparisons on this deterministic split only; they are not directly comparable to studies evaluated on the original public test-only release.

\begin{table}[!htb]
  \centering
  \singletablefont
  \setlength{\tabcolsep}{2pt}
  \renewcommand{\arraystretch}{1.04}
  \begin{tabular}{@{}l@{\hspace{4pt}}l@{\hspace{6pt}}c@{\hspace{6pt}}l@{}}
    \toprule
    \textbf{Dataset} & \textbf{Public Split} & \textbf{Final Split} & \textbf{Tracks} \\
    \midrule
    STATE-ToxiCN & official train/test & 6424 / 1605 & ZH-main, ZH-quadruple \\
    TBO & public test only & 3200 / 800 & EN-main \\
    \bottomrule
  \end{tabular}
  \caption{Datasets used in the experiments. TBO is repartitioned from its public test set with \texttt{seed=20260415}.}
  \label{tab:data_summary}
\end{table}

\paragraph{Evaluation metrics}
The original evaluation protocols of both benchmarks are kept. No mixed cross-lingual total score is constructed. For Chinese, following STATE-ToxiCN, both Hard and Soft Macro-F1 are reported: Hard requires exact agreement with gold spans and field combinations, while Soft credits overlapping spans under the same target-centred structure \cite{bai-etal-2025-state}.

On the Chinese tracks, Target and Argument evaluate target and argument field parsing. T-A Pair requires the target to be correctly bound to its argument. T-A-H Tri and Quad require the local structure to remain correct after adding the hateful label and, in the four-field task, the \texttt{group} field. For English, the paper follows TBO's tuple-level evaluation: Target and Argument measure field-level parsing; Targeted requires joint parsing of target and harmfulness; NTA and TA require exact match on $(target, argument)$ and $(target, argument, harmful)$ respectively; Harm evaluates harmfulness on the predicted target tuples \cite{zampieri-etal-2023-target}. Chinese metrics therefore emphasise hard/soft field parsing, whereas English metrics emphasise exact tuple consistency \cite{bai-etal-2025-state,zampieri-etal-2023-target}.

\subsection{Experimental Setup and Baselines}
\label{subsec:setup}

\paragraph{Experimental setup}
The main local experiments use Qwen3-14B under a dual-24GB-class GPU budget \cite{yang2025qwen3technicalreport}. Phase~1--3 share the same backbone, with lexicon injection used only for the Chinese Cultural Bystander. Long runs support checkpoint-resume and OOM split-retry. Phase~4 is deterministic aggregation. Sample-level outputs follow the concatenate-first strategy of Section~\ref{subsec:phase4}. Appendix~\ref{sec:appendix_repro} gives the runtime configuration and artifact map; Appendix~\ref{sec:appendix_auditor} gives the arbitration settings.

\paragraph{Baselines}
The baselines cover direct extraction, task-adapted SynChain and Dance, and SPAR with local and API backbones \cite{fan-etal-2025-aspect,wang_danceha_2026,deepseek_v4_2026,openai2026introducinggpt54}. Table~\ref{tab:api_zeroshot} reports the added full-test API zero-shot averages. Prompt templates and baseline adaptations appear in Appendices~\ref{sec:appendix_prompts} and~\ref{sec:appendix_baselines}.

\begin{table}[!htb]
  \centering
  \singletablefont
  \setlength{\tabcolsep}{5pt}
  \begin{tabular}{lcc}
    \toprule
    \textbf{API zero-shot model} & \textbf{EN} & \textbf{ZH} \\
    \midrule
    DeepSeek-v4-Pro & \best{23.18} & 22.80 \\
    GPT-5.4 & 21.01 & \best{28.13} \\
    \bottomrule
  \end{tabular}
  \caption{Bold marks the highest score in each panel or metric column. Full-test average scores for the added API zero-shot baselines.}
  \label{tab:api_zeroshot}
\end{table}

\begin{table*}[!t]
  \centering
  \small
  \setlength{\tabcolsep}{3pt}
  \renewcommand{\arraystretch}{1}

  \begin{minipage}{\textwidth}
    \centering

    \textbf{(a) \texttt{ZH-main}}\par
    \vspace{1pt}
    \begin{tabular}{@{\hspace{5pt}}lccccccccc@{\hspace{5pt}}}
      \toprule
      \textbf{Model} & \multicolumn{2}{c}{\textbf{Target}} & \multicolumn{2}{c}{\textbf{Argument}} & \multicolumn{2}{c}{\textbf{T-A Pair}} & \multicolumn{2}{c}{\textbf{T-A-H Tri.}} & \textbf{Avg.} \\
      \cmidrule(lr){2-3} \cmidrule(lr){4-5} \cmidrule(lr){6-7} \cmidrule(lr){8-9}
       & \textbf{Hard} & \textbf{Soft} & \textbf{Hard} & \textbf{Soft} & \textbf{Hard} & \textbf{Soft} & \textbf{Hard} & \textbf{Soft} & \\
      \midrule
      \multicolumn{10}{@{}l}{\textit{Local LLMs}} \\
    0-shot-Qwen3-14B & 45.49 & \second{54.97} & 16.24 & 46.99 & 10.62 & 32.73 & 7.65 & 23.93 & 29.83 \\
    SPAR-Qwen3-14B (mainline) & \best{48.73} & \best{55.75} & \best{21.03} & \second{56.85} & \best{12.36} & \second{34.14} & \second{8.71} & \second{24.97} & \best{32.82} \\
    SPAR-Qwen3.5-35B & \second{45.84} & 53.12 & \second{20.09} & \best{57.95} & \second{11.23} & \best{35.87} & \best{9.26} & \best{27.24} & \second{32.58} \\
      \midrule
      \multicolumn{10}{@{}l}{\textit{API Models}} \\
    few-shot-DeepSeek-v4-Pro & 52.69 & 63.08 & 17.86 & 61.03 & 13.62 & 39.53 & 10.75 & 31.09 & 36.21 \\
    few-shot-GPT-5.4 & \second{54.17} & \second{65.83} & 18.87 & \best{62.10} & 13.82 & 43.90 & 10.60 & 31.47 & 37.60 \\
    SynChain-DeepSeek-v4-Pro & 38.92 & 48.68 & 11.54 & 49.57 & 5.09 & 31.89 & 3.75 & 23.75 & 26.65 \\
    SynChain-GPT-5.4 & 34.89 & 43.15 & 11.52 & 45.63 & 5.38 & 30.28 & 4.04 & 21.98 & 24.61 \\
    Dance-DeepSeek-v4-Pro & 53.19 & 62.77 & 19.34 & 57.40 & 15.75 & 41.52 & 12.55 & 32.47 & 36.87 \\
    Dance-GPT-5.4 & 51.99 & 60.73 & \second{23.89} & 57.72 & \best{19.32} & \second{44.56} & \best{13.63} & \second{33.99} & \second{38.23} \\
    SPAR-DeepSeek-v4-Pro & \best{56.80} & \best{65.96} & \best{24.03} & \second{62.06} & 17.27 & 44.13 & 13.01 & 32.88 & \best{39.52} \\
    SPAR-GPT-5.4 & 52.62 & 61.94 & 23.12 & 56.62 & \second{17.42} & \best{45.48} & \second{13.32} & \best{34.38} & 38.11 \\
      \bottomrule
    \end{tabular}

    \vspace{4pt}

    \textbf{(b) \texttt{EN-main}}\par
    \vspace{1pt}
    \begin{tabular}{@{\hspace{5pt}}lccccccc@{\hspace{5pt}}}
      \toprule
      \textbf{Model} & \textbf{Target} & \textbf{Argument} & \textbf{Targeted} & \textbf{NTA} & \textbf{TA} & \textbf{Harm} & \textbf{Avg.} \\
      \midrule
      \multicolumn{8}{@{}l}{\textit{Local LLMs}} \\
    0-shot-Qwen3-14B & 32.41 & 46.48 & 25.06 & 14.91 & 11.06 & 47.61 & 29.59 \\
    SPAR-Qwen3-14B (mainline) & \second{43.45} & \second{47.96} & \second{34.68} & \best{20.96} & \best{18.68} & \second{59.31} & \second{37.51} \\
    SPAR-Qwen3.5-35B & \best{46.49} & \best{50.22} & \best{35.43} & \second{20.28} & \second{16.72} & \best{60.11} & \best{38.21} \\
      \midrule
      \multicolumn{8}{@{}l}{\textit{API Models}} \\
    few-shot-DeepSeek-v4-Pro & 36.26 & 42.40 & 22.12 & 3.09 & 1.73 & 53.98 & 26.60 \\
    few-shot-GPT-5.4 & 38.94 & 44.02 & 13.50 & 4.00 & 2.61 & 48.16 & 25.21 \\
    SynChain-DeepSeek-v4-Pro & 38.70 & 41.22 & 18.26 & 2.85 & 2.58 & \best{58.00} & 26.94 \\
    SynChain-GPT-5.4 & 36.60 & 37.52 & 15.56 & 4.76 & 3.52 & 51.92 & 24.98 \\
    Dance-DeepSeek-v4-Pro & 45.67 & 36.08 & 27.43 & 2.92 & 2.61 & 56.44 & 28.53 \\
    Dance-GPT-5.4 & \second{46.15} & 42.39 & 24.92 & 8.88 & 7.75 & 52.62 & 30.45 \\
    SPAR-DeepSeek-v4-Pro & \best{51.31} & \best{50.36} & \best{30.02} & \best{19.81} & \second{13.33} & 56.19 & \best{36.84} \\
    SPAR-GPT-5.4 & 42.15 & \second{47.00} & \second{29.24} & \second{18.84} & \best{15.81} & \second{56.78} & \second{34.97} \\
      \bottomrule
    \end{tabular}
  \end{minipage}

  \caption{Main results on \texttt{ZH-main} and \texttt{EN-main}. Within each split, the best result in each column is boldfaced and the second-best is underlined.}
  \label{tab:main_bilingual}
\end{table*}
\subsection{Main Results}
\label{subsec:main_results}

Table~\ref{tab:main_bilingual} reports the complete bilingual main-track comparison.

\paragraph{Main bilingual triplet results}
Under a fixed local 14B backbone, SPAR raises the average score on \texttt{ZH-main} from $29.83$ to $32.82$ and on \texttt{EN-main} from $29.59$ to $37.51$. The larger gains occur on stricter joint structural metrics.

Qwen3.5-35B records $38.21$ on \texttt{EN-main} and $32.58$ on \texttt{ZH-main}. Among API systems, \texttt{SPAR-DeepSeek-v4-Pro} records the highest \texttt{ZH-main} average ($39.52$), while \texttt{SPAR-GPT-5.4} records EN NTA/TA scores of $18.84/15.81$.

\paragraph{Chinese quadruple extension}
The Chinese four-field extension (\texttt{ZH-quadruple}) retains the \texttt{group} field as a test of adaptation to language-specific schema. Full results appear in Appendix Table~\ref{tab:appendix_main_zh_quad}.

\paragraph{Cross-model observations}
The largest differences occur on Targeted, NTA, and TA for English and on T-A Pair, T-A-H Tri, and Quad for Chinese. Appendix~\ref{sec:appendix_qualitative} provides additional outputs and cases.

\subsection{Framework Analysis}
\label{subsec:analysis}

\subsubsection*{Controlled Comparisons}

\textbf{Integrated-prompt controls.} Integrated prompting yields lower strict joint scores, although its Target scores exceed SPAR for GPT-5.4 (Table \ref{tab:integrated_control}). The controls fix the Phase 1 focus units and combine all three perspective descriptions in one extraction call, retaining the same lexicon access on Chinese. They remove separate candidate outputs, clustering, and Auditor arbitration. The comparison therefore evaluates the combined organisation of these operations.

\begin{table}[!htb]
\centering
\singletablefont
\setlength{\tabcolsep}{3pt}
\textbf{(a) English NTA / TA}\par
\begin{tabular}{@{}lcc@{}}
\toprule
Backbone & Integrated & SPAR \\
\midrule
DeepSeek-v4-Pro & 4.87 / 3.89 & \textbf{19.81} / \textbf{13.33} \\
GPT-5.4 & 11.41 / 8.18 & \textbf{18.84} / \textbf{15.81} \\
\bottomrule
\end{tabular}
\par\medskip
\textbf{(b) Chinese Pair Hard / Tri Hard}\par
\begin{tabular}{@{}lcc@{}}
\toprule
Backbone & Integrated & SPAR \\
\midrule
DeepSeek-v4-Pro & 6.31 / 4.40 & \textbf{17.27} / \textbf{13.01} \\
GPT-5.4 & 14.08 / 10.56 & \textbf{17.42} / \textbf{13.32} \\
\bottomrule
\end{tabular}
\caption{Full-test fixed-Phase-1 integrated-prompt controls. The integrated condition retains the three lens descriptions and Chinese lexicon access, but removes separate candidates, clustering, and Auditor arbitration. Bold marks the higher value for each metric within a backbone.}
\label{tab:integrated_control}
\end{table}

\textbf{Repeated-sampling control.} The three-role condition calls each role prompt once, while repeated sampling calls the same generic extraction prompt independently three times on the same focus input. Each call can return multiple candidate tuples. The reported Qwen3-14B, DeepSeek-v4-Pro, and GPT-5.4 settings retain the preceding input processing, clustering, Auditor, and sample-level reassembly (Tables \ref{tab:repeat_body_zh} and \ref{tab:repeat_body_en}).

Role-conditioned prompting scores higher on all six English metrics and Chinese Hard joint metrics across the three tested backbones. Chinese Soft results depend on the backbone, with GPT-5.4 and Qwen3-14B improving on T-A Pair Soft and T-A-H Tri Soft and DeepSeek scoring lower. This comparison supports the tested role-prompt family within the retained pipeline.

\begin{table}[!htb]
\centering
\singletablefont
\setlength{\tabcolsep}{3pt}
\begin{tabular}{@{}lcccc@{}}
\toprule
Condition & \multicolumn{2}{c}{T-A Pair} & \multicolumn{2}{c}{T-A-H Tri.} \\
 & Hard & Soft & Hard & Soft \\
\midrule
\multicolumn{5}{@{}l}{\textit{DeepSeek-v4-Pro}} \\
Three roles & \textbf{18.27} & 48.54 & \textbf{13.61} & 35.44 \\
Repeated sampling & 13.58 & \textbf{54.69} & 7.32 & \textbf{37.60} \\
\midrule
\multicolumn{5}{@{}l}{\textit{GPT-5.4}} \\
Three roles & \textbf{17.42} & \textbf{45.48} & \textbf{13.32} & \textbf{34.38} \\
Repeated sampling & 13.71 & 39.48 & 13.03 & 30.81 \\
\midrule
\multicolumn{5}{@{}l}{\textit{Qwen3-14B}} \\
Three roles & \textbf{12.36} & \textbf{34.14} & \textbf{8.71} & \textbf{24.97} \\
Repeated sampling & 7.94 & 27.61 & 7.24 & 22.64 \\
\bottomrule
\end{tabular}
\caption{Repeated-sampling control on Chinese. Scores are percentages. T-A Pair and T-A-H Tri. Hard/Soft Macro-F1 are shown. Bold marks the higher score within each backbone. Complete metrics appear in Appendix~\ref{sec:appendix_repeat}.}
\label{tab:repeat_body_zh}
\end{table}

\begin{table}[!htb]
\centering
\singletablefont
\setlength{\tabcolsep}{3pt}
\begin{tabular}{@{}lccc@{}}
\toprule
Condition & Targeted & NTA & TA \\
\midrule
\multicolumn{4}{@{}l}{\textit{DeepSeek-v4-Pro}} \\
Three roles & \textbf{32.03} & \textbf{21.10} & \textbf{13.91} \\
Repeated sampling & 19.56 & 7.33 & 3.14 \\
\midrule
\multicolumn{4}{@{}l}{\textit{GPT-5.4}} \\
Three roles & \textbf{29.24} & \textbf{18.84} & \textbf{15.81} \\
Repeated sampling & 21.34 & 9.97 & 7.35 \\
\midrule
\multicolumn{4}{@{}l}{\textit{Qwen3-14B}} \\
Three roles & \textbf{34.68} & \textbf{20.96} & \textbf{18.68} \\
Repeated sampling & 27.62 & 12.07 & 11.64 \\
\bottomrule
\end{tabular}
\caption{Repeated-sampling control on English. Scores are percentages. Targeted, NTA, and TA are shown. Bold marks the higher score within each backbone. Complete metrics appear in Appendix~\ref{sec:appendix_repeat}.}
\label{tab:repeat_body_en}
\end{table}

\subsubsection*{Component Ablations}

\textbf{Phase ablations.} On the Qwen3-14B mainline, removing the segmenter raises Target scores in both languages but lowers Chinese Hard joint scores and English TA and Harm. Removing arbitration lowers average performance on both tracks (Table \ref{tab:ablation_summary}). These changes distinguish gains in target identification from losses in structural recovery.

\textbf{Role ablations.} Removing the Cultural Bystander slightly raises the Chinese average while lowering T-A Pair Hard and T-A-H Tri Hard. Removing the Moderator raises English NTA and lowers TA. The average therefore masks weaker Chinese Hard joint recovery, and the English results distinguish two-field from three-field recovery. Complete results appear in Appendix \ref{sec:appendix_ablation}.

\begin{table}[!htb]

  \centering
  \singletablefont
  \setlength{\tabcolsep}{2pt}
  \renewcommand{\arraystretch}{1.06}
  \textbf{(a) \texttt{ZH-main}}\par
  \vspace{2pt}
  \begin{tabular}{@{}>{\raggedright\arraybackslash}p{0.36\columnwidth}ccccc@{}}
    \toprule
    \textbf{Setting} & \textbf{Tgt-H} & \textbf{Arg-H} & \textbf{TA-H} & \textbf{TAH-H} & \textbf{Avg.} \\
    \midrule
    Qwen3-14B (mainline) & 48.73 & \best{21.03} & \best{12.36} & \best{8.71} & 32.82 \\
    w/o P1 segmenter & \best{52.34} & 15.20 & 9.26 & 7.02 & 32.25 \\
    w/o P2 victim & 50.40 & 15.95 & 10.53 & 7.63 & 31.16 \\
    w/o P2 moderator & 49.90 & 16.51 & 10.86 & 7.75 & 31.77 \\
    w/o P2 bystander & 50.18 & 16.10 & 10.50 & 8.20 & \best{32.85} \\
    w/o P3 arbitration & 46.54 & 17.03 & 10.67 & 7.68 & 31.71 \\
    \bottomrule
  \end{tabular}

  \vspace{4pt}
  \textbf{(b) \texttt{EN-main}}\par
  \vspace{2pt}
  \begin{tabular}{@{}>{\raggedright\arraybackslash}p{0.36\columnwidth}ccccc@{}}
    \toprule
    \textbf{Setting} & \textbf{Tgt} & \textbf{NTA} & \textbf{TA} & \textbf{Harm} & \textbf{Avg.} \\
    \midrule
    Qwen3-14B (mainline) & 43.45 & 20.96 & \best{18.68} & \best{59.31} & \best{37.51} \\
    w/o P1 segmenter & \best{46.95} & 20.65 & 18.04 & 57.21 & 37.34 \\
    w/o P2 victim & 44.70 & 21.18 & 17.20 & 57.21 & 37.02 \\
    w/o P2 moderator & 44.29 & \best{21.34} & 17.51 & 59.11 & 37.43 \\
    w/o P3 arbitration & 43.66 & 20.42 & 17.81 & 57.61 & 36.72 \\
    \bottomrule
  \end{tabular}
  \caption{Bold marks the highest score in each panel or metric column. Ablation summary on the bilingual main tracks. The main text keeps only the most diagnostic hard/exact metrics; complete bilingual ablation results appear in Appendix~\ref{sec:appendix_ablation}.}
  \label{tab:ablation_summary}

\end{table}

\subsection{Distillation Results}
\label{subsec:distill}

Table~\ref{tab:distill_summary} compares the distilled 4B student with the zero-shot and SPAR-Qwen3-14B systems on all three evaluation tracks.

\begin{table}[!htb]
  \centering
  \singletablefont
  \setlength{\tabcolsep}{4pt}
  \textbf{(a) \texttt{ZH-main}}\par
  \vspace{2pt}
  \begin{tabular}{lccc}
    \toprule
    \textbf{Model} & \textbf{TAH-H} & \textbf{TAH-S} & \textbf{Avg.} \\
    \midrule
    0-shot-Qwen3-14B & 7.65 & 23.93 & 29.83 \\
    SPAR-Qwen3-14B (mainline) & \best{8.71} & 24.97 & 32.82 \\
    Distillation-Qwen3-4B & 8.33 & \best{26.04} & \best{33.67} \\
    \bottomrule
  \end{tabular}

  \vspace{4pt}
  \textbf{(b) \texttt{EN-main}}\par
  \vspace{2pt}
  \begin{tabular}{lccc}
    \toprule
    \textbf{Model} & \textbf{TA} & \textbf{Harm} & \textbf{Avg.} \\
    \midrule
    0-shot-Qwen3-14B & 11.06 & 47.61 & 29.59 \\
    SPAR-Qwen3-14B (mainline) & \best{18.68} & \best{59.31} & \best{37.51} \\
    Distillation-Qwen3-4B & 18.50 & 59.05 & 37.33 \\
    \bottomrule
  \end{tabular}

  \vspace{4pt}
  \textbf{(c) \texttt{ZH-quadruple}}\par
  \vspace{2pt}
  \begin{tabular}{lccc}
    \toprule
    \textbf{Model} & \textbf{Quad-H} & \textbf{Quad-S} & \textbf{Avg.} \\
    \midrule
    0-shot-Qwen3-14B & 6.87 & 19.42 & 25.70 \\
    SPAR-Qwen3-14B (mainline) & 6.18 & 22.37 & 28.95 \\
    Distillation-Qwen3-4B & \best{7.04} & \best{25.05} & \best{30.38} \\
    \bottomrule
  \end{tabular}
  \caption{Bold marks the highest score in each panel or metric column. Distillation summary across the three evaluation tracks.}
  \label{tab:distill_summary}
\end{table}

The 4B student trained on teacher traces scores higher on average than SPAR-Qwen3-14B on both Chinese tracks and remains close to the teacher pipeline on the English average. Appendix~\ref{sec:appendix_distill} reports teacher-trace construction, filtering, and student training.

\subsection{Qualitative Analysis}
\label{subsec:qualitative}

\paragraph{Case study}
Appendix Table~\ref{tab:case_zh} traces three Phase~1 local focus units from a Chinese multi-target example. Phase~2 produces conflicting labels for $g_2$ and $g_3$, and Phase~3 selects the gold label for both units. For $g_1$, SPAR retains the full argument ``poor; populous'', which SynChain truncates.

\paragraph{Error analysis}
The main residual error is argument boundary over-expansion. In Appendix Table~\ref{tab:error_spar}, SPAR-Hate predicts the correct target and harmfulness but retains additional contextual material, so the prediction receives no tuple credit under exact match. Appendix~\ref{sec:appendix_qualitative} provides further cases.

\FloatBarrier
\section{Conclusion}

SPAR-Hate improves joint target--argument--label recovery on the controlled Chinese and English benchmarks. Integrated-prompt comparisons favour separate generation and arbitration, while repeated-sampling comparisons favour role-conditioned generation under exact matching. Chinese overlap-based comparisons vary by backbone. Teacher traces support a 4B student whose average scores exceed the 14B teacher pipeline on both Chinese tracks and remain close on English.

\section*{Limitations}
\label{sec:limitations}

Evaluation covers Chinese and English and omits a full comparison with task-specific fine-tuned systems. The task-motivated role set has not been exhaustively searched, and the lexicon has no complete removal ablation. Weak lexical context can still over-flag dialectal, culturally loaded, or reclaimed terms. The threshold replay, blind audit, and attacked-group breakdown have limited scope and cannot support broad claims about statistical optimality, fairness, or cultural validity. TBO results use the deterministic internal split and do not support direct comparison with results on the original public test-only release.

\section*{Ethical Considerations}

This study uses public Chinese and English benchmark datasets to examine structured hate speech parsing. Their contents include abusive, discriminatory, and potentially traumatic expressions. The paper retains only representative examples needed for the academic argument and recommends a minimal-exposure principle in data processing, visualisation, and manual analysis to reduce secondary harm to researchers, annotators, and readers.

Such systems also carry misuse risks, including over-censorship, automated punishment, and large-scale opinion monitoring. Earlier work shows that false positives can arise from ambiguous boundaries between offensive language and hate speech, spurious correlations around minority-group mentions, and model fragility on functional phenomena \cite{davidson_automated_2017,hartvigsen-etal-2022-toxigen,rottger-etal-2021-hatecheck}. Structured outputs and traceable intermediate processes increase transparency, but they may also increase the operational reach of deployed moderation systems. Real-world deployment therefore requires human review, appeals, threshold calibration, error auditing, and continuing bias evaluation across languages, groups, and cultural settings.

\bibliography{Spar_references}

\appendix
\raggedbottom
\makeatletter
\setlength{\@fptop}{0pt}
\setlength{\@fpsep}{12pt}
\setlength{\@fpbot}{0pt plus 1fil}
\setlength{\@dblfptop}{0pt}
\setlength{\@dblfpsep}{12pt}
\setlength{\@dblfpbot}{0pt plus 1fil}
\makeatother
\setlength{\textfloatsep}{8pt plus 1pt minus 1pt}
\setlength{\floatsep}{8pt plus 1pt minus 1pt}
\setlength{\intextsep}{8pt plus 1pt minus 1pt}

\section{Implementation and Artifact Details}
\label{sec:appendix_repro}

Table~\ref{tab:appendix_runtime} gives the retained \texttt{Qwen3-14B} configuration \cite{yang2025qwen3technicalreport}. Phase~1--3 share the same local backbone, and Phase~4 is deterministic. The compact profile uses \texttt{vLLM} \cite{kwon2023efficientmemorymanagementlarge}.

\begin{table}[htbp]
  \centering
  \singletablefont
  \setlength{\tabcolsep}{4pt}
  \begin{tabular}{p{0.33\columnwidth}p{0.57\columnwidth}}
    \toprule
    \textbf{Item} & \textbf{Value} \\
    \midrule
    Backbone & Qwen3-14B \\
    Hardware & RTX 3090 + RTX 4090D \\
    Precision & \texttt{bfloat16} \\
    Context length & 8192 \\
    Tensor parallelism & 2 \\
    Phase~1 & local focus-unit segmentation; batch 4 \\
    Phase~2 & 3 roles; micro-batch 64; Chinese bystander with lexicon injection \\
    Phase~3 & quality threshold 0.60; cluster similarity 0.66; auditor enabled; max retained clusters 2 \\
    Runtime safeguards & checkpoint resume; OOM split-retry \\
    Compact service profile & \texttt{vLLM} backend; auditor batch 1; regeneration disabled \\
    Phase~4 & deterministic reassembly \\
    \bottomrule
  \end{tabular}
  \caption{Main local configuration for the reported Qwen3-14B mainline.}
  \label{tab:appendix_runtime}
\end{table}

\FloatBarrier

\subsection{Runtime and Call Diagnostics}

Retained EN, ZH-main, and ZH-quadruple runs take 499.0, 617.6, and 623.3 minutes. Phase~2 makes 4,524, 6,978, and 6,933 role calls at 0.25--0.28 role-groups per second. Auditor refinement is triggered for 982, 1,645, and 1,610 groups. These figures describe retained runs; incomplete billing logs preclude a monetary cost estimate.

\subsection{Artifact Map and Recoverability}

The artifact roots \texttt{SPAR/}, \texttt{SPAR\_anonymous\_submission/}, and \texttt{transfer\_bundles/} contain the deterministic TBO split and Phase~0 preparation path, Phase~1--3 prompts and configurations, Phase~3 scoring and fallback code, direct and API baseline outputs, SynChain and Dance adaptations, diagnostic summaries, and teacher/student exports. Historical bilingual distillation records are partly archival. Preserved reports and official evaluation exports support the reported construction and evaluation statistics, while the Chinese student bundle retains the independently recoverable training example.

\section{Prompt Templates}
\label{sec:appendix_prompts}

The three Phase~2 roles share the output contract, Target Resolution, Non-empty Argument, Output Sequence, Short Rationale, and Current Task blocks. Figure~\ref{fig:appendix_prompt_phase2_victim} gives the Victim instruction body and example headers. The Moderator and Cultural Bystander figures retain their role-specific additions. Repeated example titles are omitted.

\begin{figure}[htbp]
  \centering
  \promptpaneltitle{Phase~1 Segmenter}
  \promptbox{%
  \fontsize{7.0}{8.0}\selectfont
  \textbf{\# Role}\\
  You are the SPAR-Hate Segment Agent specializing in English text.\\
  Your ONLY output is a strict JSON array.\\[2pt]
  \textbf{\# Rules}\\
  1. Divide by target and intent: Split the text into separate groups if different targets are attacked or distinct intents exist.\\
  2. Context isolation: \texttt{group\_text} MUST be copied EXACTLY from the original text as a meaningful local context. Do not paraphrase, summarize, or alter punctuation/emojis.\\
  3. Target normalization: \texttt{canonical\_target} is the direct entity mention. If the target is strictly absent, unmentioned, or null, output ``\texttt{IMPLICIT\_TARGET}''.\\
  4. Output scope: ONLY output \texttt{local\_group\_id}, \texttt{group\_text}, and \texttt{canonical\_target}.\\[2pt]
  \textbf{\# Examples}\\
  Example 1 (Multi-Target Segmentation)\\
  Example 2 (Null/Implicit Targets \& General Venting)\\
  Example 3 (Non-harmful / Object Target with Emojis)\\[2pt]
  \textbf{\# Current Task}\\
  Original text: \texttt{\{\{TEXT\}\}}\\
  Output: \texttt{[}%
  }
  \caption{Excerpt of the English Phase~1 segmenter prompt used in the bilingual main tracks.}
  \label{fig:appendix_prompt_phase1}
\end{figure}

\begin{figure}[htbp]
  \centering
  \promptpaneltitle{Victim}
  \promptbox{%
  \fontsize{7.0}{8.0}\selectfont
  \textbf{\# Role}\\
  You are the ``Victim Persona'' Agent in a Hate Speech Evaluation system specializing in English text.\\
  You are highly sensitive, empathetic to marginalized groups, and acutely aware of emotional harm, exclusion, stereotyping, slurs, and microaggressions.\\
  Your ONLY output is a strict JSON array.\\[2pt]
  \textbf{\# Rules}\\
  1. Victim's Lens: Put yourself in the shoes of the attacked group. If the text inside the focus area makes you feel unsafe, dehumanized, degraded, or stereotyped, flag it as harmful (\texttt{harmful: 1}).\\
  2. Attention Lock (CRITICAL): Your evaluation area is STRICTLY limited to the text inside the \texttt{<focus>} and \texttt{</focus>} tags. You MUST NOT penalize offensive behaviors or slurs that exist outside these tags.\\
  3. Target Resolution: prioritize the target inside \texttt{<focus>}; use outside context or \texttt{Target hint} only when the focus text is fragmentary; do not override an explicit in-focus target. If the focus only contains an insult or adjective but no addressee, output ``\texttt{IMPLICIT\_TARGET}''.\\
  4. Argument Extraction: \texttt{argument} MUST be an exact substring from INSIDE the \texttt{<focus>} tags and should be the shortest decisive span that carries the insult, stereotype, exclusion, or offensive action.\\
  5. Multi-Argument Extraction (CRITICAL): Split independent attacks into separate objects. Do NOT merge two insults or predicates into one long argument just because they are joined by and, \&, commas, or nearby context.\\
  6. Non-empty Argument (CRITICAL): \texttt{argument} must NEVER be empty. Even for harmless outputs (\texttt{harmful: 0}), choose the shortest exact substring inside \texttt{<focus>} that best supports the harmless judgment.\\
  7. Output Sequence: each object must strictly follow \texttt{rationale}, \texttt{target}, \texttt{argument}, \texttt{harmful}.\\
  8. Short Rationale: \texttt{rationale} must be brief (under 30 words), single-line only, and must not contain raw newline characters inside the JSON string.\\[2pt]
  \textbf{\# Examples}\\
  Example 1 (Missing Subject Resolved via Context)\\
  Example 2 (Multiple Arguments on the Same Target within Focus)\\
  Example 3 (Multi-Argument Extraction on Permissible/Non-Harmful Text)\\[2pt]
  \textbf{\# Current Task}\\
  Target hint (soft only; use only when focus lacks a subject): \texttt{"\{\{TARGET\_HINT\}\}"}\\
  Input: \texttt{"\{\{FOCUS\_TEXT\}\}"}\\
  Output: \texttt{[}%
  }
  \caption{Main instruction body of the English Victim prompt used in Phase~2.}
  \label{fig:appendix_prompt_phase2_victim}
\end{figure}

\begin{figure}[htbp]
  \centering
  \promptpaneltitle{Moderator}
  \promptbox{%
  \fontsize{7.0}{8.0}\selectfont
  \textbf{\# Role}\\
  You are the ``Moderator Persona'' Agent in a Hate Speech Evaluation system specializing in English text.\\
  Your duty is to enforce social media platform Terms of Service (TOS) and community safety guidelines. You look for actionable offenses: severe slurs, targeted harassment, incitement of violence, and hate speech. You tolerate general profanity and subjective disagreements if they do not cross into targeted abuse.\\[2pt]
  \textbf{\# Role-specific rules}\\
  1. Moderator's Lens: Evaluate if the text INSIDE the focus area violates safety policies. Explicit slurs, hostile degradation, threats of violence, or targeted abuse must be flagged as harmful (\texttt{harmful: 1}).\\
  4. Argument Extraction: \texttt{argument} should be the shortest decisive span that contains the actionable abuse, slur, threat, or hostile phrase.\\[2pt]
  \textbf{\# Example headers}\\
  Example 1 (Severe Violation with Pronoun Resolution)\\
  Example 2 (Explicit Slurs and Threats)\\
  Example 3 (Harmless/No Violation inside Focus)%
  }
  \caption{Role-specific English Moderator prompt used in Phase~2. Shared blocks with Victim are omitted.}
  \label{fig:appendix_prompt_phase2_moderator}
\end{figure}

\begin{figure}[h]
  \centering
  \promptpaneltitle{Cultural Bystander}
  \promptbox{%
  \fontsize{7.0}{8.0}\selectfont
  \textbf{\# Role}\\
  You are the ``Cultural Bystander Persona'' Agent in a Hate Speech Evaluation system specializing in English text.\\
  You act as an internet anthropologist. You focus on the sociological nuances of language: distinguishing reclaimed slang from derogatory usage, identifying group generalization templates, recognizing coded derogation, and differentiating between quoting/irony and genuine hostility. You assess the speaker's stance and tone.\\[2pt]
  \textbf{\# Role-specific rules}\\
  1. Cultural Lens: Evaluate the societal and cultural implications of the text INSIDE the focus area. Recognize when a slur is reclaimed by the community versus when it is weaponized. Flag generalizations and coded derogation as harmful if they enforce negative societal stereotypes.\\
  Treat reclaimed slang as non-harmful only when the focus itself shows clear affiliative or celebratory usage. Negative complaints, commands, mockery, or group generalizations are not reclaimed evidence.\\
  4. Argument Extraction: \texttt{argument} should be the shortest decisive span that captures the key cultural attack, slang, stereotype, or behavior.\\
  5. Multi-Target/Argument Extraction (CRITICAL): Split independent attacks into separate objects.\\[2pt]
  \textbf{\# Example headers}\\
  Example 1 (Reclaimed Slang \& Performative Tone)\\
  Example 2 (Group Generalization)%
  }
  \caption{Role-specific English Cultural Bystander prompt used in Phase~2. Shared blocks with Victim are omitted.}
  \label{fig:appendix_prompt_phase2_bystander}
\end{figure}
\begin{figure}[h]
  \centering
  \promptpaneltitle{Few-shot API Baseline}
  \promptbox{%
  \fontsize{7.0}{8.0}\selectfont
  You are a content review expert. Please identify hate speech in the sentence and extract one or more triplets.\\
  Your ONLY output is a strict JSON array.\\
  Each object must contain: \texttt{rationale}, \texttt{target}, \texttt{argument}, \texttt{harmful}.\\
  \texttt{target}: the discussed or attacked entity; if there is no explicit target, output ``\texttt{IMPLICIT\_TARGET}''.\\
  \texttt{argument}: the core evidence span.\\
  \texttt{harmful}: must be \texttt{1} or \texttt{0}.\\
  Output JSON array only. Do not output any explanation.\\[2pt]
  Positive example: ``The men where I live are trash.''\\
  Output: \texttt{(target=men,\ argument=trash,\ harmful=1)}\\[2pt]
  Negative example: ``What kind of dogs are the ugly gross ones with bloodshot eyes and long ass ears''\\
  Output: \texttt{(target=dogs,\ argument=ugly\ gross\ ones\ with\ bloodshot\ eyes\ and\ long\ ass\ ears,\ harmful=0)}\\[2pt]
  Input: \texttt{"\{\{TEXT\}\}"}\\
  Output: return the answer starting exactly with \texttt{[} and ending with \texttt{]}%
  }
  \caption{English few-shot baseline prompt shared by the DeepSeek-v4-Pro and GPT-5.4 API baselines.}
  \label{fig:appendix_prompt_fewshot}
\end{figure}
\begin{figure}[h]
  \centering
  \promptpaneltitle{Phase~3 Auditor}
  \promptbox{%
  \fontsize{7.0}{8.0}\selectfont
  \textbf{\#\# Role}\\
  You are the SPAR-Hate Phase-3 Arbitrate Auditor for English. You do not perform crude majority vote. You arbitrate between role evidence, candidate clusters, triage signals, and dynamic beacons to select the final tuple most likely to match the English ground truth.\\
  Your ONLY output is a strict JSON object.\\[2pt]
  \textbf{\#\# English Task Rules}\\
  1. Ground the final answer in \texttt{focus\_text}. \texttt{final\_argument} must be an exact or near-exact substring from the focus span.\\
  2. Prefer \texttt{final\_target} from inside the focus span. Only use \texttt{canonical\_target} or outside context when the focus span is fragmentary, pronominal, or missing the subject.\\
  5. Distinguish reclaimed slang, quotation, refutation, irony, and performative speech from genuine targeted harm.\\
  7. If one role misses coded derogation or over-penalizes a reclaimed or quoted expression, say so compactly in \texttt{fused\_rationale} and \texttt{correction\_trace}.\\
  9. If several candidates differ mainly in argument length, prefer the shortest span that still independently expresses the attack core.\\[2pt]
  \textbf{\#\# Few-shot example headers}\\
  Example 1: clear group generalization, final harmful\\
  Example 2: reclaimed slang, final non-harmful\\
  Example 3: shrink an over-broad argument span%
  }
  \caption{Rendered excerpt of the English Phase~3 auditor prompt used in the mainline configuration.}
  \label{fig:appendix_prompt_phase3}
\end{figure}

\section{Baseline Adaptation}
\label{sec:appendix_baselines}

SynChain and Dance are modified only enough to produce hate tuples compatible with the benchmarks. The aim is to preserve each method's high-level inductive bias rather than rewrite it into a new multi-stage system \cite{fan-etal-2025-aspect,wang_danceha_2026}.

\paragraph{General adaptation.}
Across all baselines, the adaptation layer makes only three minimal changes. First, raw outputs are normalised into strict JSON readable by the official evaluation scripts. Second, field names and label spaces are aligned: English uses \texttt{harmful}, Chinese uses \texttt{hateful}, and \texttt{group} is enabled only on the Chinese four-field track. Third, lightweight post-processing is applied only where required by the evaluation interface, including boundary cleaning, duplicate tuple removal, and unified sample-level export. None of these operations adds candidate arbitration or reranking.

\paragraph{SynChain.}
The syntax-aware extraction bias is retained. The adapted workflow still generates structure-sensitive candidates first, then resolves target, argument, and label, and finally maps intermediate candidates back to scorable text fields. Beyond alignment to the output contract, no extra multi-role generation, auditor arbitration, or lexicon weighting is added.

\paragraph{Dance.}
The divide-and-conquer skeleton of grouping, per-group inference, and merge is retained. The adapted system still performs inference over local groups before returning to sample-level outputs. The only change is that the internal extraction target is rewritten from aspect/opinion/sentiment-style fields into hate tuples. As with SynChain, no extra SPAR-style arbitration or lexicon-aware reranking is added.

\section{Auditor Scoring and Bounded Arbitration}
\label{sec:appendix_auditor}

Phase~3 scores, clusters, routes, and selects role candidates under grounding constraints. Table~\ref{tab:appendix_beacons} gives the main settings, and Table~\ref{tab:appendix_scoring_features} gives the scoring signals.

\begin{table}[H]
  \centering
  \singletablefont
  \setlength{\tabcolsep}{3pt}
  \begin{tabular}{p{0.31\columnwidth}p{0.61\columnwidth}}
    \toprule
    \textbf{Item} & \textbf{Value} \\
    \midrule
    Role set & victim / moderator / cultural\_bystander \\
    Quality threshold & 0.60 \\
    Cluster similarity & 0.66 \\
    Similarity weights & target/argument/label/group = 0.45/0.45/0.05/0.05 \\
    Score bonuses & support role 0.12; candidate lexicon 0.05; cluster lexicon 0.08 \\
    Auditor lanes & consensus / defective / conflict / lexicon-hit / victim-microaggression \\
    Auditor batch & 2 (main 14B); 1 (compact \texttt{vllm}) \\
    Regeneration & enabled (main 14B); disabled (compact \texttt{vllm}) \\
    Cluster retention & max 2; score floor 1.05; min support roles 2 \\
    Decoding profile & \texttt{bfloat16}; TP=2; context 8192 \\
    Final selection & deterministic fallback after optional refinement \\
    \bottomrule
  \end{tabular}
  \caption{Key Phase~3 settings in the local 14B arbitration configurations.}
  \label{tab:appendix_beacons}
\end{table}

\begin{table}[H]
  \centering
  \singletablefont
  \setlength{\tabcolsep}{3pt}
  \begin{tabular}{p{0.34\columnwidth}p{0.58\columnwidth}}
    \toprule
    \textbf{Feature group} & \textbf{Signal} \\
    \midrule
    Grounding & argument in focus text; no invented evidence \\
    Target anchoring & explicit target; canonical-target match \\
    Structural validity & JSON validity; legal label; track-compatible fields \\
    Group-label coherence & \texttt{group}--label consistency in \texttt{ZH-quadruple} \\
    Soft priors & lexicon bonus; moderator floor; victim microaggression \\
    \bottomrule
  \end{tabular}
  \caption{Main feature groups used by the Phase~3 candidate scorer.}
  \label{tab:appendix_scoring_features}
\end{table}

The quality threshold of 0.60 filters weak or ungrounded candidates before clustering. The similarity threshold of 0.66 limits merges between candidates with divergent \texttt{target} or \texttt{argument} fields. Both values are reused across the reported languages and benchmarks.

\begin{table}[H]
  \centering
  \singletablefont
  \setlength{\tabcolsep}{3pt}
  \begin{tabular}{p{0.18\columnwidth}p{0.51\columnwidth}p{0.21\columnwidth}}
    \toprule
    \textbf{Step} & \textbf{Operation} & \textbf{Bound} \\
    \midrule
    Validate & Check JSON, label, argument grounding, and target anchor & quality $\geq 0.60$ \\
    Cluster & Compare target, argument, label, and optional group & similarity $\geq 0.66$ \\
    Route & Assign consensus, defective, or conflict lane & one local unit \\
    Refine & Apply soft beacons and optional auditor call & one regeneration \\
    Select & Rank with deterministic fallback & at most two clusters \\
    \bottomrule
  \end{tabular}
  \caption{Operational summary of local Phase~3 arbitration.}
  \label{tab:appendix_auditor_steps}
\end{table}

\begin{table}[H]
  \centering
  \singletablefont
  \setlength{\tabcolsep}{3pt}
  \begin{tabular}{p{0.46\columnwidth}p{0.44\columnwidth}}
    \toprule
    \textbf{Measure} & \textbf{Retained-run summary} \\
    \midrule
    Raw-cluster p95 & EN 3; ZH 4 \\
    Retained/final p95 and maximum & 2 \\
    Conflict-lane share & 55.9--61.7\% \\
    Auditor-use share & 65.1--70.7\% \\
    \bottomrule
  \end{tabular}
  \caption{Boundedness and routing diagnostics across the retained runs.}
  \label{tab:appendix_auditor_diagnostics}
\end{table}

\section{Robustness and Semantic Validation}
\label{sec:appendix_validation}

\subsection{Threshold Sensitivity}

The actual auditor is replayed on fixed seed-10947 subsets of 200 complete samples per language, covering 385 EN and 281 ZH focus groups per setting. Quality varies over $0.55/0.60/0.65$ at similarity $0.66$. Similarity varies over $0.60/0.66/0.72$ at quality $0.60$. All other settings remain fixed. Table~\ref{tab:appendix_threshold_replay} reports the retained summary.

\begin{table}[htbp]
  \centering
  \singletablefont
  \setlength{\tabcolsep}{4pt}
  \begin{tabular}{lcc}
    \toprule
    \textbf{Language} & \textbf{Observed range} & \textbf{Default} \\
    \midrule
    EN & 34.97--35.08 & 35.08 \\
    ZH & 26.28--27.12 & 26.82 \\
    \bottomrule
  \end{tabular}
  \caption{Four-hard-metric averages in the fixed-subset threshold replay. The results are not full-test significance estimates.}
  \label{tab:appendix_threshold_replay}
\end{table}

\subsection{Blind Semantic Audit}

Two independent bilingual or native-speaker reviewers assess 60 stratified Chinese focus units with gold labels hidden. Separate final-tuple and Cultural-Bystander-rationale rubrics yield 120 judgements. Table~\ref{tab:appendix_blind_audit} gives acceptance and agreement. Consensus cases reach 70\% strict overall acceptance; disagreement-heavy lexicon--auditor cases receive lower acceptance.

\begin{table}[htbp]
  \centering
  \singletablefont
  \setlength{\tabcolsep}{3pt}
  \begin{tabular}{lcccc}
    \toprule
    \textbf{Object} & \textbf{R1} & \textbf{R2} & \textbf{Agree.} & $\boldsymbol{\kappa}$ \\
    \midrule
    Final tuple & 73.3 & 71.7 & 81.7 & 0.540 \\
    Bystander rationale & 68.3 & 58.3 & 86.7 & 0.716 \\
    \bottomrule
  \end{tabular}
  \caption{Acceptance rates, percentage agreement, and Cohen's $\kappa$ in the blind semantic audit.}
  \label{tab:appendix_blind_audit}
\end{table}

\subsection{Strict Sample Exact-Set Summary}

Sample exact-set matching requires the complete predicted tuple set to match the gold set. A missing tuple, an extra tuple, or a mismatch in any field makes the sample fail this criterion.

\begin{table}[htbp]
  \centering
  \singletablefont
  \setlength{\tabcolsep}{3pt}
  \begin{tabular}{lcc}
    \toprule
    \textbf{Track} & \textbf{SPAR-DeepSeek} & \textbf{Direct references} \\
    \midrule
    EN & 20/800 (2.50\%) & 0--9/800 (0.00--1.13\%) \\
    ZH-main & 141/1605 (8.79\%) & zero-shot: 37/1605; 66/1605 \\
    \bottomrule
  \end{tabular}
  \caption{Counts and percentages of samples whose predicted tuple sets exactly match the gold sets.}
  \label{tab:appendix_exact_set}
\end{table}

SPAR-DeepSeek has a higher complete-sample match rate than the direct references on English and the two zero-shot references on Chinese, while some Chinese few-shot conditions score higher. Chinese joint tuple scores and complete-sample recovery rank the systems differently, and the low complete-sample match rates indicate remaining difficulty in recovering whole-sample outputs.

\subsection{Attacked-Group Breakdown}

Table~\ref{tab:appendix_group_diagnostic} reports ZH-quadruple scores by gold attacked group. The Racism+Sexism group has the largest TAH-to-Quad drop, locating the main loss in multi-group binding. The breakdown measures robustness across the annotated groups; fairness requires separate evidence.

\begin{table}[htbp]
  \centering
  \singletablefont
  \setlength{\tabcolsep}{3pt}
  \begin{tabular}{lrrr}
    \toprule
    \textbf{Gold group} & \textbf{Samples} & \textbf{TAH-S} & \textbf{Quad-S} \\
    \midrule
    Sexism & 302 & 36.66 & 34.49 \\
    Racism & 223 & 40.56 & 35.57 \\
    Region & 216 & 27.20 & 24.93 \\
    LGBTQ & 93 & 34.75 & 32.13 \\
    Racism + Sexism & 85 & 21.85 & 5.30 \\
    \bottomrule
  \end{tabular}
  \caption{ZH-quadruple attacked-group diagnostic. Scores are Soft F1 and groups are defined from gold labels.}
  \label{tab:appendix_group_diagnostic}
\end{table}

\section{Qualitative Cases and Error Analysis}
\label{sec:appendix_qualitative}

\begin{table}[!htb]
  \centering
  \singletablefont
  \setlength{\tabcolsep}{2pt}
  \renewcommand{\arraystretch}{1.06}
  \begin{tabular}{@{}>{\raggedright\arraybackslash}p{0.17\columnwidth}>{\raggedright\arraybackslash}p{0.77\columnwidth}@{}}
    \toprule
    \textbf{Item} & \textbf{Content} \\
    \midrule
    Text &
    \parbox[t]{\linewidth}{First impression: Henan is poor and populous; Northeast China has one of the world's lowest birth rates; public security in Xinjiang may not be very good.} \\
    Gold &
    \parbox[t]{\linewidth}{$g_1$: (Henan, ``poor; populous'', 1)\\
    $g_2$: (Northeast China, one of the world's lowest birth rates, 1)\\
    $g_3$: (Xinjiang, public security may not be very good, 1)} \\
    Phase~1 &
    \parbox[t]{\linewidth}{$g_1$: Henan: poor; populous\\
    $g_2$: Northeast China: one of the world's lowest birth rates\\
    $g_3$: Xinjiang: public security may not be very good} \\
    Phase~2 &
    \parbox[t]{\linewidth}{$g_1$: only hateful candidates remain.\\
    $g_2$: (Northeast China, one of the world's lowest birth rates, 1/0).\\
    $g_3$: (Xinjiang, public security may not be very good, 1/0).} \\
    Phase~3 &
    \parbox[t]{\linewidth}{$g_1$: (Henan, ``poor; populous'', 1)\\
    $g_2$: (Northeast China, one of the world's lowest birth rates, 1)\\
    $g_3$: (Xinjiang, public security may not be very good, 1)} \\
    Baselines &
    \parbox[t]{\linewidth}{$g_1$: Dance is correct. SynChain outputs (Henan, poor, 0), missing ``populous''.\\
    $g_2$: Dance and SynChain both assign label 0.\\
    $g_3$: Dance and SynChain both assign label 0.} \\
    \bottomrule
  \end{tabular}
  \caption{Phase-wise processing of three local focus units. $g_1$, $g_2$, and $g_3$ denote Phase~1 units. One unit may yield several sample-level tuples after Phase~4. The Chinese input is shown in English translation.}
  \label{tab:case_zh}
\end{table}

\begin{table}[!htb]
  \centering
  \singletablefont
  \setlength{\tabcolsep}{6pt}
  \begin{tabular}{@{}p{1.8cm}p{5.4cm}@{}}
    \toprule
    \textbf{Item} & \textbf{Content} \\
    \midrule
    Text & some of y'all aren't being held accountable for the dumb shit that y'all be doing....and it shows \\
    Gold & (y'all, dumb shit, 1) \\
    SPAR-Qwen3-14B & (y'all, dumb shit that y'all be doing, 1); (null, it shows, 0) \\
    \bottomrule
  \end{tabular}
  \caption{A remaining SPAR-Hate error: overly wide argument boundaries.}
  \label{tab:error_spar}
\end{table}

The Chinese case isolates a target conflict. The English case traces a conflict-lane decision within a multi-target sample.

\begin{table}[h]
  \centering
  \singletablefont
  \setlength{\tabcolsep}{3pt}
  \renewcommand{\arraystretch}{1.04}
  \begin{tabular}{@{}p{0.29\columnwidth}p{0.63\columnwidth}@{}}
    \toprule
    \textbf{Field} & \textbf{Content} \\
    \midrule
    Text & The people spreading the claim that Henan people steal manhole covers are in fact Beijingers. \\
    Gold tuple & (Beijingers, saying that Henan people steal manhole covers, 1) \\
    Victim & (Beijingers, saying that Henan people steal manhole covers, 1) \\
    Moderator & (Henan people, steal manhole covers, 1) \\
    Cultural Bystander & (Beijingers, saying that Henan people steal manhole covers, 1) \\
    Triage & \texttt{lane=conflict}; \texttt{initial\_conflict=target\_conflict} \\
    Beacons & \texttt{grounding}; \texttt{consensus} \\
    Correction trace & \texttt{moderator: downweighted for misreading the victim mention as the attacking target} \\
    Final tuple & (Beijingers, saying that Henan people steal manhole covers, 1) \\
    \bottomrule
  \end{tabular}
  \caption{Chinese case study showing target-conflict resolution in Phase~3.}
  \label{tab:appendix_case_zh}
\end{table}

\noindent In this case, Victim and Cultural Bystander both identify the attacking target as ``Beijingers'', whereas Moderator shifts the target to the mentioned victim group. The auditor follows the grounded consensus on the focus text and downweights the moderator reading.

\begin{table}[h]
  \centering
  \singletablefont
  \setlength{\tabcolsep}{3pt}
  \renewcommand{\arraystretch}{1.04}
  \begin{tabular}{@{}p{0.29\columnwidth}p{0.63\columnwidth}@{}}
    \toprule
    \textbf{Field} & \textbf{Content} \\
    \midrule
    Text &
    \textit{that shit is not cute \& dnt make u hard.} \\
    Phase~4 restored sample prediction &
    \parbox[t]{0.63\columnwidth}{\texttt{(bitches, Ghetto, 1)}; \texttt{(bitches, ignorant, 1)};\\
    \texttt{(shit, shit, 1)}; \texttt{(IMPLICIT\_TARGET, alldat shit, 0)}} \\
    Victim &
    \parbox[t]{0.63\columnwidth}{\texttt{(shit, shit, 1)}\\
    \texttt{(u, dnt make u hard, 1)}} \\
    Moderator &
    \parbox[t]{0.63\columnwidth}{\texttt{(IMPLICIT\_TARGET, that shit, 0)}\\
    \texttt{(IMPLICIT\_TARGET, dnt make u hard, 0)}} \\
    Cultural Bystander &
    \texttt{(shit, shit, 0)} \\
    Triage & \texttt{lane=conflict} \\
    Beacons & \texttt{grounding}, \texttt{victim\_microaggression} \\
    Correction trace & \texttt{moderator: downweighted}; \texttt{cultural\_bystander: downweighted} \\
    Final tuple & \texttt{(shit, shit, 1)} \\
    \bottomrule
  \end{tabular}
  \caption{English case study showing conflict-lane arbitration inside a multi-target sample.}
  \label{tab:appendix_case_en}
\end{table}

\noindent The current text is focus unit $g_2$ in a sample containing $g_1$, $g_2$, and $g_3$. Phase~4 retains two harmful tuples from $g_1$, \texttt{(shit, shit, 1)} from $g_2$, and one non-harmful tuple from $g_3$. One focus unit can emit several tuples. For $g_2$, the auditor selects the Victim reading and downweights the non-harmful Moderator and Cultural Bystander candidates.

\section{Distillation Data and Student Training}
\label{sec:appendix_distill}

\subsection{Teacher-trace Schema}

\begin{table}[h]
  \centering
  \singletablefont
  \setlength{\tabcolsep}{3pt}
  \renewcommand{\arraystretch}{1.08}
  \begin{tabular}{p{0.36\columnwidth}p{0.56\columnwidth}}
    \toprule
    \textbf{Field group} & \textbf{Content} \\
    \midrule
    Identifiers & \texttt{sample\_id}, \texttt{group\_id}, source, split, language \\
    Source text & original text and the current focus/group unit \\
    Focus evidence & \texttt{group\_text}, \texttt{focus\_text}, \texttt{canonical\_target} \\
    Role evidence & tuples from the three roles, short rationales, quality scores, lexicon hit \\
    Conflict diagnosis & triage lane, field conflicts, applied beacons, clusters \\
    Teacher decision & \texttt{teacher\_final\_tuple}, \texttt{teacher\_rationale}, audit confidence \\
    Supervision messages & system instruction, user evidence package, assistant supervision target \\
    \bottomrule
  \end{tabular}
  \caption{Schema of the final constructed distillation rows.}
  \label{tab:appendix_trace_schema}
\end{table}

The distillation target includes more than the final tuple. Each training row also preserves intermediate decomposition, role candidates, conflict diagnosis, and the final arbitration result, so that the student learns a structured decision process rather than only the flat output surface.

\subsection{Filtering and Dataset Statistics}

\begin{table}[h]
  \centering
  \singletablefont
  \setlength{\tabcolsep}{3pt}
  \renewcommand{\arraystretch}{1.02}
  \begin{tabular}{p{0.34\columnwidth}p{0.14\columnwidth}p{0.38\columnwidth}}
    \toprule
    \textbf{Stage} & \textbf{Count} & \textbf{Note} \\
    \midrule
    Phase~3 group traces & 7,752 & all available local decision traces \\
    Packaged train rows & 3,739 & base samples entering the teacher package \\
    Teacher reasoning HQ & 2,221 & main warm-up variant \\
    HQ Chinese rows & 1,590 & \texttt{zh\_group\_hateful} \\
    HQ English rows & 631 & \texttt{en\_span\_harmful} \\
    Teacher reasoning plus & 3,194 & reasoning variant with relaxed conflict threshold \\
    Teacher JSON-only plus & 3,194 & final JSON-aligned variant \\
    Validation checks & all pass & duplicate id 0; schema error 0; invalid final tuple 0 \\
    \bottomrule
  \end{tabular}
  \caption{Summary statistics for the teacher-trace construction pipeline.}
  \label{tab:appendix_trace_stats}
\end{table}

The bilingual statistics come from the main distillation construction run. The pipeline has three steps: first, retain train rows with recoverable identifiers and serialisable teacher packages; second, build a strict HQ reasoning subset using groundedness, final tuple validity, and audit confidence; third, expand it under the same grounding constraints into reasoning-plus and JSON-only variants for curriculum training.

\subsection{A Real Constructed Training Example}

\begin{table}[htbp]
  \centering
  \singletablefont
  \setlength{\tabcolsep}{3pt}
  \renewcommand{\arraystretch}{1.04}
  \begin{tabular}{@{}p{0.28\columnwidth}p{0.64\columnwidth}@{}}
    \toprule
    \textbf{Field} & \textbf{Content} \\
    \midrule
    Source text &
    Marriage is a necessity for women. Men can stay unmarried for life, but women cannot. In reality, countless men are even more eager. They will empty the family savings and act like simps just to marry and have children. Absurd. \\
    Sample-level gold &
    (marriage, a necessity for women, 1); (countless men, empty the family savings and act like simps just to marry and have children, 1) \\
    Current focus &
    In reality, countless men are even more eager. They will empty the family savings and act like simps just to marry and have children. Absurd. \\
    Dataset variant &
    \texttt{teacher\_trace\_reasoning\_hq} \\
    Victim &
    tuple: (men, even more eager than women, and willing to empty the family savings and act like simps just to marry and have children, harmful); rationale: It describes men as simps and derides their behaviour, carrying clear gender stereotyping and disparagement. \\
    Moderator &
    tuple: (men, willing to empty the family savings and act like simps just to marry and have children, harmful); rationale: The focus uses derogatory wording such as ``simps'' to negatively characterise men. The phrasing is attacking and demeaning, which violates community standards. \\
    Cultural Bystander &
    tuple: (men, act like simps, harmful); rationale: [Lexicon Hit] ``simp'' is used here to belittle men and suggest a lack of dignity. \\
    Conflict diagnosis &
    lane=\texttt{consensus}; high-risk field=\texttt{argument}; error tags=\texttt{lexicon\_hit}, \texttt{auditor\_span\_refined} \\
    Teacher final tuple &
    target=men; argument=act like simps; group=unknown; harmful=1; direction=harmful \\
    Assistant target &
    format: reasoning+JSON; final JSON: target=men; argument=act like simps; group=unknown; harmful=1; direction=harmful \\
    \bottomrule
  \end{tabular}
  \caption{A condensed real training row from the constructed distillation data.}
  \label{tab:appendix_distill_example}
\end{table}

Table~\ref{tab:appendix_distill_example} shows one real final training row from the released Chinese student bundle. The sample comes from a multi-target Chinese input, but the current training row supervises only the second focus unit. The three role rationales are retained so that the student learns argument anchoring when targets agree but arguments compete.

\subsection{Student Training Setup}

\begin{center}
  \centering
  \small
  \setlength{\tabcolsep}{4pt}
  \begin{tabular}{p{0.35\columnwidth}p{0.55\columnwidth}}
    \toprule
    \textbf{Item} & \textbf{Value} \\
    \midrule
    Student backbone & Qwen3-4B-Instruct-2507 \\
    Prompt template & Qwen3 no-think \\
    Training split & train only \\
    Supervision regime & teacher-only \\
    Curriculum stages & teacher reasoning HQ; teacher reasoning plus; teacher JSON-only plus \\
    Epochs per stage & 1.0 / 1.0 / 1.0 \\
    Sequence length & 4096 \\
    Per-device batch size & 1 \\
    Gradient accumulation & 8 \\
    Optimizer and schedule & learning rate 1e-4; cosine; warmup ratio 0.05 \\
    Precision & bf16 \\
    Adaptation method & QLoRA (4-bit BnB) \\
    LoRA setting & rank 32; alpha 32; dropout 0.0 \\
    \bottomrule
  \end{tabular}
  \captionof{table}{Student-model training setup used in the distillation pipeline.}
  \label{tab:appendix_student_setup}
\end{center}

Student training follows a teacher-only curriculum. Stage~1 uses the strict HQ reasoning subset to establish the structured decision process. Stage~2 expands coverage with reasoning-plus. Stage~3 aligns the model to strict JSON output. Under this protocol, training uses only the train split, ground truth is never used as assistant supervision, evaluation is limited to teacher reasoning and teacher JSON, and manual prefill is disallowed. The \texttt{Qwen3-4B} backbone, 4-bit QLoRA, and LoRA rank/alpha settings follow the corresponding parameter-efficient fine-tuning methods \cite{yang2025qwen3technicalreport,dettmers2023qloraefficientfinetuningquantized,hu2021lora}.

\section{Repeated-Sampling Protocol and Full Results}
\label{sec:appendix_repeat}
Tables~\ref{tab:repeat_full_zh} and~\ref{tab:repeat_full_en} report the full metrics for the comparison summarised in Section~\ref{subsec:analysis}. Each condition uses three initial generation calls on the same focus inputs, and each call may return multiple tuples. The comparison retains the clustering, Auditor, and sample-level reassembly pipeline. Matching initial call counts does not equate total token use or downstream arbitration cost.

The DeepSeek summaries belong to the repeated-sampling comparison series, whereas Table~\ref{tab:main_bilingual} retains the earlier main-result series. Absolute scores are compared within each series. Chinese Target Hard and Soft values are reproduced as reported. The aggregate summaries do not support paired confidence intervals or significance estimates.
\begin{table}[H]
\centering\singletablefont
\setlength{\tabcolsep}{1.5pt}
\textbf{(a) Field metrics}\par
\begin{tabular*}{\columnwidth}{@{\extracolsep{\fill}}lcccc@{}}
\toprule
Condition & \multicolumn{2}{c}{Target} & \multicolumn{2}{c}{Argument} \\
 & Hard & Soft & Hard & Soft \\
\midrule
\multicolumn{5}{@{}l}{\textit{DeepSeek-v4-Pro}} \\
Three roles & \textbf{61.22} & 58.76 & \textbf{27.79} & \textbf{67.97} \\
Repeated & 47.56 & \textbf{74.36} & 18.32 & 64.47 \\
\midrule
\multicolumn{5}{@{}l}{\textit{GPT-5.4}} \\
Three roles & \textbf{52.62} & \textbf{61.94} & \textbf{23.12} & \textbf{56.62} \\
Repeated & 45.32 & 56.53 & 20.20 & 52.21 \\
\midrule
\multicolumn{5}{@{}l}{\textit{Qwen3-14B}} \\
Three roles & \textbf{48.73} & \textbf{55.75} & \textbf{21.03} & \textbf{56.85} \\
Repeated & 41.01 & 51.76 & 18.78 & 53.27 \\
\bottomrule
\end{tabular*}
\par\medskip
\textbf{(b) Joint metrics}\par
\begin{tabular*}{\columnwidth}{@{\extracolsep{\fill}}lcccc@{}}
\toprule
Condition & \multicolumn{2}{c}{T-A Pair} & \multicolumn{2}{c}{T-A-H Tri.} \\
 & Hard & Soft & Hard & Soft \\
\midrule
\multicolumn{5}{@{}l}{\textit{DeepSeek-v4-Pro}} \\
Three roles & \textbf{18.27} & 48.54 & \textbf{13.61} & 35.44 \\
Repeated & 13.58 & \textbf{54.69} & 7.32 & \textbf{37.60} \\
\midrule
\multicolumn{5}{@{}l}{\textit{GPT-5.4}} \\
Three roles & \textbf{17.42} & \textbf{45.48} & \textbf{13.32} & \textbf{34.38} \\
Repeated & 13.71 & 39.48 & 13.03 & 30.81 \\
\midrule
\multicolumn{5}{@{}l}{\textit{Qwen3-14B}} \\
Three roles & \textbf{12.36} & \textbf{34.14} & \textbf{8.71} & \textbf{24.97} \\
Repeated & 7.94 & 27.61 & 7.24 & 22.64 \\
\bottomrule
\end{tabular*}
\caption{Complete repeated-sampling results on Chinese. Scores are percentages. Repeated denotes repeated sampling. Bold marks the higher score within each backbone.}
\label{tab:repeat_full_zh}
\end{table}

\clearpage
\begin{table*}[p]
\begin{minipage}{\columnwidth}
\centering\singletablefont
\setlength{\tabcolsep}{1.5pt}
\begin{tabular*}{\columnwidth}{@{\extracolsep{\fill}}lcccccc@{}}
\toprule
Condition & Target & Argument & Targeted & NTA & TA & Harm \\
\midrule
\multicolumn{7}{@{}l}{\textit{DeepSeek-v4-Pro}} \\
Three roles & \textbf{49.75} & \textbf{49.46} & \textbf{32.03} & \textbf{21.10} & \textbf{13.91} & \textbf{57.43} \\
Repeated & 47.26 & 35.10 & 19.56 & 7.33 & 3.14 & 47.81 \\
\midrule
\multicolumn{7}{@{}l}{\textit{GPT-5.4}} \\
Three roles & \textbf{42.15} & \textbf{47.00} & \textbf{29.24} & \textbf{18.84} & \textbf{15.81} & \textbf{56.78} \\
Repeated & 40.13 & 38.44 & 21.34 & 9.97 & 7.35 & 53.13 \\
\midrule
\multicolumn{7}{@{}l}{\textit{Qwen3-14B}} \\
Three roles & \textbf{43.45} & \textbf{47.96} & \textbf{34.68} & \textbf{20.96} & \textbf{18.68} & \textbf{59.31} \\
Repeated & 41.66 & 40.81 & 27.62 & 12.07 & 11.64 & 56.56 \\
\bottomrule
\end{tabular*}
\caption{Complete repeated-sampling results on English. Scores are percentages. Repeated denotes repeated sampling. Bold marks the higher score within each backbone.}
\label{tab:repeat_full_en}
\end{minipage}
\end{table*}

\begin{table*}[p]
\begin{minipage}{\columnwidth}
\section{Full Bilingual Ablation Results}
\label{sec:appendix_ablation}
Table~\ref{tab:appendix_ablation_full} gives the complete bilingual ablations underlying Section~\ref{subsec:analysis}.
\end{minipage}\par\medskip

  \centering
  \scriptsize
  
  \setlength{\tabcolsep}{3.4pt}
  \renewcommand{\arraystretch}{0.94}
    \textbf{(a) \texttt{ZH-main}}\par
    \vspace{2pt}
    \begin{tabular}{@{\hspace{3pt}}lccccccccc@{\hspace{3pt}}}
      \toprule
      \textbf{Setting} & \multicolumn{2}{c}{\textbf{Target}} & \multicolumn{2}{c}{\textbf{Argument}} & \multicolumn{2}{c}{\textbf{T-A Pair}} & \multicolumn{2}{c}{\textbf{T-A-H Tri.}} & \textbf{Avg.} \\
      \cmidrule(lr){2-3} \cmidrule(lr){4-5} \cmidrule(lr){6-7} \cmidrule(lr){8-9}
       & \textbf{Hard} & \textbf{Soft} & \textbf{Hard} & \textbf{Soft} & \textbf{Hard} & \textbf{Soft} & \textbf{Hard} & \textbf{Soft} & \\
      \midrule
    Qwen3-14B (mainline) & 48.73 & 55.75 & \best{21.03} & \best{56.85} & \best{12.36} & 34.14 & \best{8.71} & 24.97 & 32.82 \\
    w/o phase1 segmenter & \best{52.34} & \best{64.46} & 15.20 & 48.92 & 9.26 & 34.43 & 7.02 & 26.39 & 32.25 \\
    w/o phase2 victim & 50.40 & 60.67 & 15.95 & 46.48 & 10.53 & 32.84 & 7.63 & 24.78 & 31.16 \\
    w/o phase2 moderator & 49.90 & 60.07 & 16.51 & 47.65 & 10.86 & 34.91 & 7.75 & 26.49 & 31.77 \\
    w/o phase2 bystander & 50.18 & 61.11 & 16.10 & 50.90 & 10.50 & \best{36.93} & 8.20 & \best{28.90} & \best{32.85} \\
    w/o phase3 & 46.54 & 56.64 & 17.03 & 51.42 & 10.67 & 36.58 & 7.68 & 27.10 & 31.71 \\
      \bottomrule
    \end{tabular}

    \vspace{5pt}

    \textbf{(b) \texttt{EN-main}}\par
    \vspace{2pt}
    \begin{tabular}{@{\hspace{3pt}}lccccccc@{\hspace{3pt}}}
      \toprule
      \textbf{Setting} & \textbf{Target} & \textbf{Arg.} & \textbf{Targeted} & \textbf{NTA} & \textbf{TA} & \textbf{Harm} & \textbf{Avg.} \\
      \midrule
    Qwen3-14B (mainline) & 43.45 & 47.96 & \best{34.68} & 20.96 & \best{18.68} & \best{59.31} & \best{37.51} \\
    w/o phase1 segmenter & \best{46.95} & 47.11 & 34.06 & 20.65 & 18.04 & 57.21 & 37.34 \\
    w/o phase2 victim & 44.70 & \best{49.17} & 32.68 & 21.18 & 17.20 & 57.21 & 37.02 \\
    w/o phase2 moderator & 44.29 & 48.25 & 34.07 & \best{21.34} & 17.51 & 59.11 & 37.43 \\
    w/o phase2 bystander & 44.49 & 47.24 & 33.10 & 20.64 & 18.37 & 59.22 & 37.18 \\
    w/o phase3 & 43.66 & 46.77 & 34.07 & 20.42 & 17.81 & 57.61 & 36.72 \\
      \bottomrule
    \end{tabular}
  \caption{Bold marks the highest score in each panel or metric column. Complete bilingual Qwen3-14B ablations.}
  \label{tab:appendix_ablation_full}
  \end{table*}

\begin{table*}[p]
\begin{minipage}{\columnwidth}
\section{Full Results on ZH-quadruple}
\label{sec:appendix_zh_quad}
Table~\ref{tab:appendix_main_zh_quad} reports the four-field results used for the Chinese extension.
\end{minipage}\par\medskip
\centering\scriptsize

  \setlength{\tabcolsep}{3.4pt}
  \renewcommand{\arraystretch}{0.94}
  \begin{tabular}{@{\hspace{3pt}}lccccccccccc@{\hspace{3pt}}}
    \toprule
    \textbf{Model} & \multicolumn{2}{c}{\textbf{Target}} & \multicolumn{2}{c}{\textbf{Argument}} & \multicolumn{2}{c}{\textbf{T-A Pair}} & \multicolumn{2}{c}{\textbf{T-A-H Tri.}} & \multicolumn{2}{c}{\textbf{Quad.}} & \textbf{Avg.} \\
    \cmidrule(lr){2-3} \cmidrule(lr){4-5} \cmidrule(lr){6-7} \cmidrule(lr){8-9} \cmidrule(lr){10-11}
     & \textbf{Hard} & \textbf{Soft} & \textbf{Hard} & \textbf{Soft} & \textbf{Hard} & \textbf{Soft} & \textbf{Hard} & \textbf{Soft} & \textbf{Hard} & \textbf{Soft} & \\
    \midrule
    \multicolumn{12}{@{}l}{\textit{Local LLMs}} \\
    0-shot-Qwen3-14B & 43.94 & 53.81 & 15.60 & 47.00 & 9.92 & 29.58 & 7.65 & 23.18 & \second{6.87} & 19.42 & 25.70 \\
    SPAR-Qwen3-14B (mainline) & \second{48.54} & \second{58.58} & \best{19.61} & \best{53.94} & \second{11.00} & \second{34.14} & \second{8.51} & \second{26.61} & 6.18 & \second{22.37} & \second{28.95} \\
    SPAR-Qwen3.5-35B & \best{50.81} & \best{60.84} & \second{19.57} & \second{52.05} & \best{13.64} & \best{38.50} & \best{10.59} & \best{29.94} & \best{8.68} & \best{24.48} & \best{30.91} \\
    \midrule
    \multicolumn{12}{@{}l}{\textit{API Models}} \\
    LLaMA3-70B* & 30.87 & 41.45 & 14.80 & 46.68 & 8.29 & 25.38 & 7.40 & 22.58 & 4.72 & 13.14 & 21.53 \\
    Claude-3.5-Sonnet* & 41.45 & 54.06 & 15.80 & 55.80 & 10.43 & 36.96 & 9.28 & \second{33.04} & 7.10 & 25.22 & 28.91 \\
    few-shot-DeepSeek-v4-Pro & \best{54.86} & \best{65.70} & 18.62 & \second{60.50} & 14.39 & 41.02 & 11.34 & 32.51 & 9.70 & \second{27.99} & 33.66 \\
    few-shot-GPT-5.4 & \second{52.87} & \second{64.98} & 17.09 & 60.28 & 12.43 & 42.17 & 9.45 & 31.54 & 8.15 & 27.50 & 32.65 \\
    SPAR-DeepSeek-v4-Pro & 52.66 & 61.29 & \best{23.73} & \best{65.66} & \best{16.45} & \second{43.12} & \second{12.70} & 32.25 & \second{10.82} & 27.65 & \best{34.63} \\
    SPAR-GPT-5.4 & 52.15 & 60.63 & \second{21.69} & 55.34 & \second{16.37} & \best{43.71} & \best{13.15} & \best{33.40} & \best{11.63} & \best{29.11} & \second{33.72} \\
    \bottomrule
  \end{tabular}
  \caption{Full Chinese four-field results. Best and second-best scores are marked within the local and API groups. Starred rows are historical STATE-ToxiCN values \cite{bai-etal-2025-state}.}
  \label{tab:appendix_main_zh_quad}
\end{table*}

\end{document}